\documentclass[11pt]{article}

\usepackage{microtype}
\usepackage{graphicx}
\usepackage{booktabs}
\usepackage{array}
\usepackage{amsmath}
\usepackage{amsfonts}
\usepackage{amssymb}
\usepackage{enumitem}
\usepackage{tikz}
\usetikzlibrary{positioning,decorations.pathreplacing}
\usepackage{microsoft-tech-report}
\usepackage{hyperref}

\newcommand{\add}[1]{{#1}}
\newcommand{\harl}{harnessed agentic RL}
\newcommand{\Harl}{Harnessed Agentic RL}

\hypersetup{
  colorlinks=true,
  linkcolor=microsoftblue,
  citecolor=microsoftblue,
  urlcolor=microsoftblue,
  pdftitle={Agent Lightning v1.0: Towards \Harl{}},
  pdfauthor={First Author, Second Author},
  pdfsubject={Technical report},
  pdfkeywords={technical report, latex template}
}

\techreportlabel{Agent Lightning v1.0: Towards \add{\Harl{}}}
\techreportshorttitle{Agent Lightning v1.0: Towards \add{\Harl{}}}
\techreportlogoheight{0.75cm}

\begin{document}
\thispagestyle{empty}
\enlargethispage{1.0cm}

\techreportlogos{%
  \techreportlogo{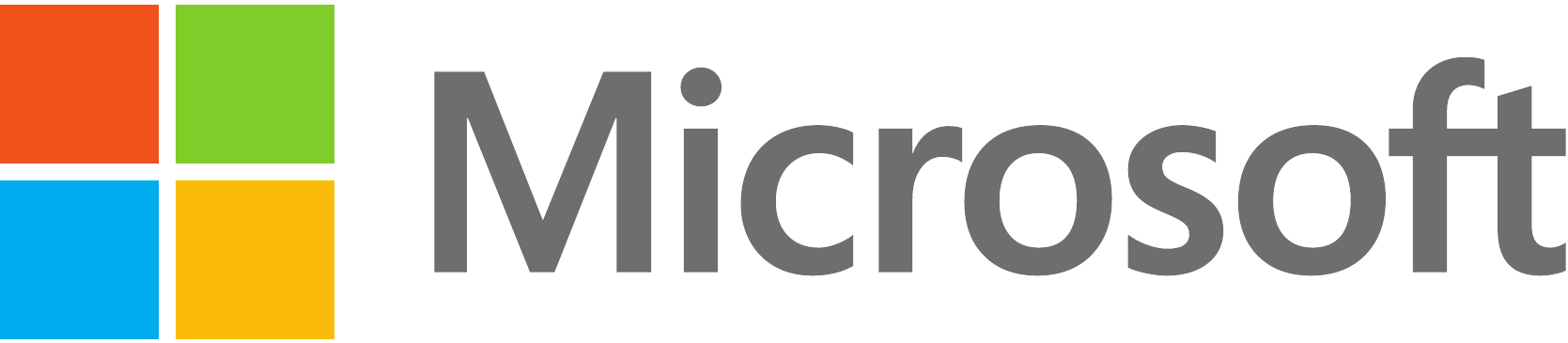}
}%
\techreportdate{\techreportmonthyear}
\techreporttoprule

\techreporttitle{\fontsize{17}{17}\selectfont Agent Lightning v1.0: Towards \add{\Harl{}}}
\techreportauthors{%
  Zhiyuan He$^{1,*,\ddagger}$ \hspace{0.7em}
  Siwei Zhang$^{2,*}$ \hspace{0.7em}
  Zhiwen Zhou$^{3,*}$ \hspace{0.7em}
  Yuqing Yang$^{1,\ddagger}$ \hspace{0.7em}
  Yu Kang$^{1}$ \hspace{0.7em} \\
  Yuge Zhang$^{1}$ \hspace{0.7em}
  Luna K. Qiu$^{1}$ \hspace{0.7em}
  Tin Yan Tsui$^{4}$ \hspace{0.7em}
  Jiahang Xu$^{1}$ \hspace{0.7em}
  Chong Luo$^{1}$
}{%
  $^{1}$Microsoft \quad
  $^{2}$Fudan University \quad
  $^{3}$Zhejiang University \quad
  $^{4}$University of Edinburgh
}

\begin{center}
\begin{minipage}{\textwidth}
  \centering
  \includegraphics[width=\textwidth]{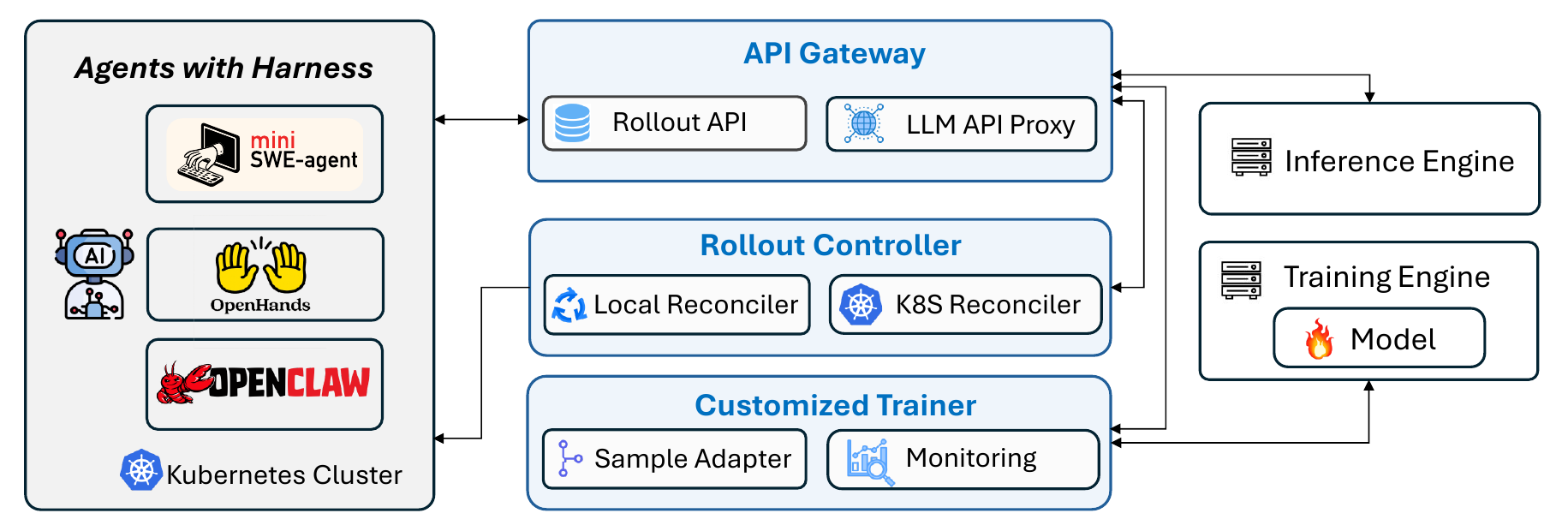}
  {\captionsetup{font=scriptsize,aboveskip=0pt,belowskip=0pt}
  \captionof{figure}{The overall framework of Agent Lightning v1.0.}
  \label{fig:teaser}}
\end{minipage}
\end{center}

\vspace{-0.2em}
\begin{microsofttitlebox}
\setlength{\parindent}{0cm}
\setlength{\parskip}{0.02cm}
\raggedright
\nohyphens

\begin{abstract}
\textbf{Abstract.}
Modern agents do not operate as standalone LLMs. They run inside \emph{agent
harnesses} that manage tools, context, and control flow, which
makes the harness a critical component. Our original Agent Lightning
work introduced a disaggregated architecture that connects arbitrary agents to
reinforcement learning (RL) training through an LLM endpoint proxy. Recent frameworks such as verl Uni-Agent, AReaL 2.0, slime v0.3.0,
and Polar have followed this proxy-based approach. Such a proxy-based training approach enables RL training with the harness. In this work, we use the
term \emph{\add{\harl{}}} to describe this paradigm, in which the
deploy-time harness is directly involved in model post-training, thereby narrowing the gap between training and actual
use.

\vspace{\baselineskip}

We find that \add{\harl{}} differs fundamentally from traditional
agentic RL and introduces a new set of challenges. In traditional agentic RL, the training engine owns the environment
interaction loop. In \add{\harl{}}, the harness owns this loop, while the training engine
observes only a sequence of LLM request-response pairs. How to model and
assemble these calls into training samples remains an open question. Through a
careful study, we identify several challenges of \harl{}, including
retokenization, sample merging, advantage calculation, loss normalization,
and training backend scheduling. We find that, if not properly addressed,
these challenges can lead to ineffective or unstable training. Existing
frameworks generally leave these issues underspecified. In this paper, we provide the first comprehensive elaboration of them.

\vspace{\baselineskip}

We further present Agent Lightning v1.0, a lightweight framework
for \add{\harl{}}. We treat simplicity as a first principle, implementing the
framework in \textbf{only approximately 3,500 lines of code.} Its compact design
supports arbitrary agent harnesses and provides a practical testbed for
studying these challenges. We validate Agent Lightning v1.0 on general
instruction-following agent, search agent, and coding agent. For coding agent, we find
that existing RL frameworks provide limited support, including a lack of
data and complete training scripts, as well as a reliance on large-scale
computational resources. To address this gap, we provide a complete
data-cleaning pipeline and reproducible training scripts based on open-source dataset and models.
\textbf{Using only 6K training examples and modest compute, RL improves
Qwen3.5-9B on SWE-bench Verified from 41.8\% to 56.4\%, an absolute 14.6\% gain.} We release the complete workflow and scripts to facilitate
reproducible \add{\harl{}} in Agent Lightning v1.0.
\vspace{\baselineskip}

\end{abstract}

\vspace{0.08cm}
{\setlength{\parskip}{0.1cm}\scriptsize
\techreportmeta{Project Page}{\href{https://github.com/microsoft/agent-lightning}{github.com/microsoft/agent-lightning}}
\techreportmeta{Correspondence}{%
\href{mailto:zhiyuhe@microsoft,yuqyang@example.com}{\{zhiyuhe, yuqyang\}@microsoft.com}}
}
\vspace{0.08cm}
{\footnotesize\rmfamily\itshape\color{microsoftgray}
$^*$Equal contribution. \quad
$^{\ddagger}$Corresponding authors.\par
}
\end{microsofttitlebox}

\clearpage
\section{Introduction}

Modern agents do not operate as standalone LLMs. They run inside \emph{agent
harnesses} that manage tools, execution environments, context, and control
flow. The harness therefore determines how an agent observes its environment,
acts over long horizons, and recovers from failures, making it a central part
of the agent's capabilities. Prominent examples include coding-agent harnesses
such as mini-SWE-agent~\cite{minisweagent}, OpenHands~\cite{openhands},
OpenCode~\cite{opencode}, Claude Code~\cite{claudecode}, and
Codex~\cite{codex}, as well as general-purpose harnesses such as
OpenClaw~\cite{openclaw} and Hermes~\cite{hermes}.

Early reinforcement learning (RL) frameworks, including verl~\cite{hybridflow},
AReaL~\cite{areal}, and slime~\cite{slime}, generally require users to
implement the agent loop directly inside the training framework. Integrating
existing agent harnesses is therefore difficult because they often have
complex implementations and their own dependencies, making them hard to integrate
directly into RL frameworks. Our original Agent Lightning
work~\cite{agentlightning} introduced a
disaggregated architecture for training and agent execution. It connects
arbitrary agents to RL training through an LLM endpoint, with almost no changes
to the agent.
More recently, such
proxy-based approach has become more common in frameworks such as verl
Uni-Agent~\cite{uniagent}, AReaL 2.0~\cite{areal2}, slime
v0.3.0~\cite{slime}, and Polar~\cite{polar}, which naturally enables enables RL training with agent harnesses. 

\add{We use the term \emph{\harl{}} for RL training conducted through
the same agent harness used at deployment. The harness, rather than the trainer, owns
context construction, tool execution, and the agent--environment interaction
loop, while the training system observes and optimizes the resulting model
calls across a service boundary. This formulation preserves the harness's
deployment-time context policy, tool protocols, and execution semantics
without requiring its agent loop to be reimplemented inside the RL framework.}

\begin{figure*}[t]
  \centering
  \begin{minipage}[c]{0.55\textwidth}
    \centering
    \includegraphics[width=\linewidth]
      {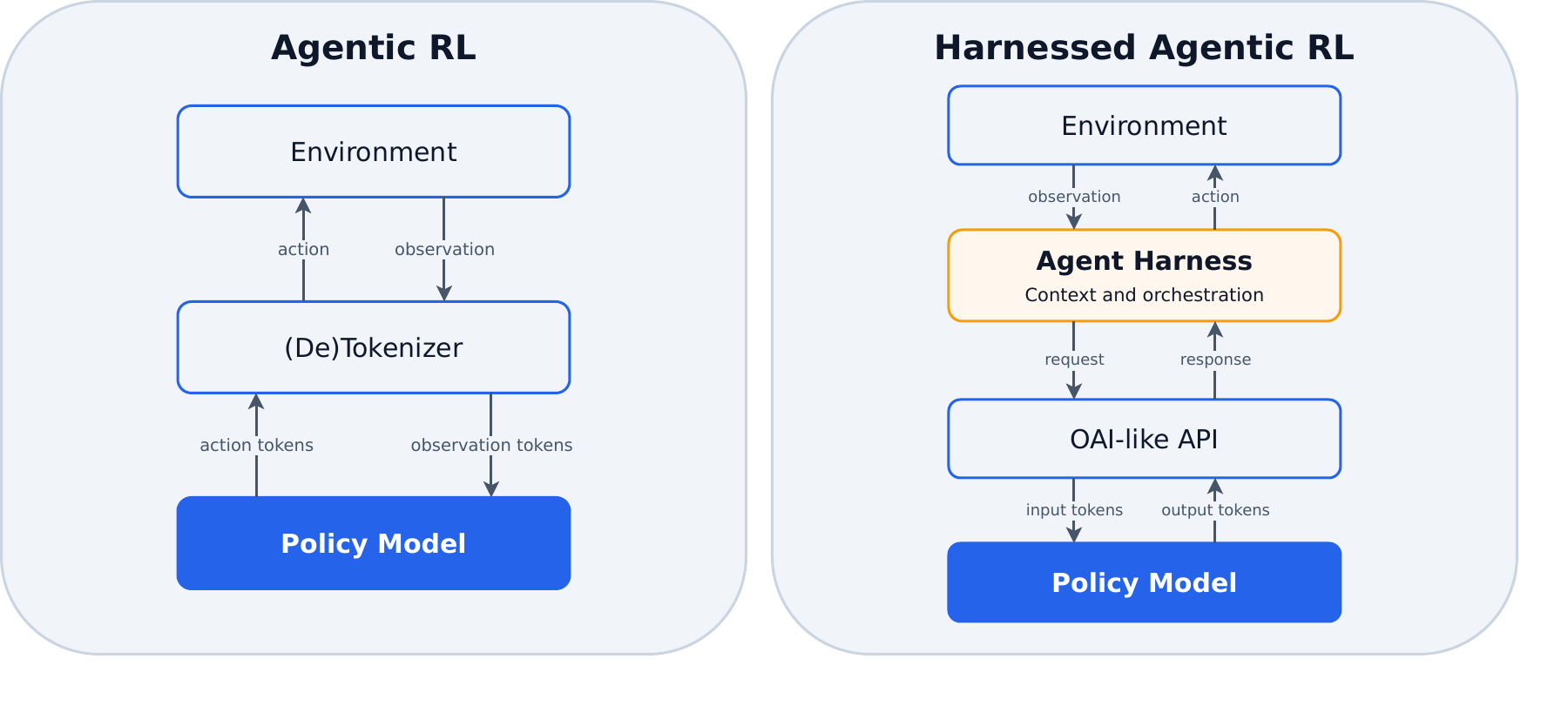}
  \end{minipage}\hfill
  \begin{minipage}[c]{0.43\textwidth}
    \centering
    \scriptsize
    \setlength{\tabcolsep}{2pt}
    \renewcommand{\arraystretch}{1.06}
    \begin{tabular}{@{}
      >{\raggedright\arraybackslash}p{0.16\linewidth}
      >{\raggedright\arraybackslash}p{0.33\linewidth}
      >{\raggedright\arraybackslash}p{0.45\linewidth}@{}}
      \toprule
      & \textbf{Agentic RL} & \textbf{\add{\Harl{}}} \\
      \midrule
      \textbf{State} &
        Environment &
        Harness + environment \\
      \hline
      \textbf{Model input} &
        Continuous token history &
        Per-call prompts \\
      \hline
      \textbf{Agents} &
        Single ReAct agent &
        Multi-agent, subagents, and handoffs \\
      \bottomrule
    \end{tabular}
  \end{minipage}
  \caption{\add{Comparison of traditional agentic RL and \harl{}.
  Both admit a POMDP formulation, but \harl{} adds harness state to
  the latent execution state and exposes the policy model to separately
  constructed model-call prompts. This change also shifts control and
  orchestration into the harness and makes the number of training samples more dynamic.}}
  \label{fig:agentic-harness-centric-comparison}
\end{figure*}

\add{Both traditional agentic RL and \harl{} can be modeled as
partially observable Markov decision processes, but they differ in their
latent state and in the observations presented to the policy model. In
traditional agentic RL, the latent state is primarily the environment state.
The policy model interacts almost directly with the environment through a transparent layer. The model
produces action tokens, the environment returns an observation, and the
tokenized observation extends the existing history as
$p_t = \bigl(p_{t-1},a_{t-1},o_t\bigr)$.
Here, $p_t$ is the token history presented to the model at step $t$,
$a_{t-1}$ is the action generated at the previous step, and $o_t$ is the
latest environment observation.
Consequently, the policy observes one continuously extended token history,
and a rollout naturally forms one linear token trajectory.

In \harl{}, the policy model no longer interacts directly with the
environment. The latent state contains both the harness state and the
environment state. The harness owns context construction, control flow, tool
execution, and agent orchestration, and independently constructs the request
prompt for each model call. The policy observes only the exact prompt delivered
through an LLM API and generates a response conditioned on that
prompt. A rollout is therefore exposed at the model boundary as a sequence of
request--response pairs,
\[
  (p_1,a_1),(p_2,a_2),\ldots,
\]
where $p_i$ is the prompt sent in the $i$-th LLM call and $a_i$ is the
corresponding model response. The intervening harness and environment state
transitions remain latent. Figure~\ref{fig:agentic-harness-centric-comparison} summarizes this
difference. It also gives rise to the implementation challenges below, for
which existing frameworks make different choices that can affect algorithmic
correctness and training stability.}

\textbf{The first challenge is retokenization and sample merging.} Agent harnesses usually
communicate with model APIs through text messages, while RL training operates
on tokens. Most frameworks merge two consecutive calls when $p_{i+1}$ contains
$(p_i,a_i)$ as a complete prefix at the token level. However, after
retokenization, even when the text is unchanged, the token IDs of $a_i$ in $p_{i+1}$ can differ from those
originally sampled by the model. This breaks
token-level continuity and prevents the two calls from being safely merged.

\textbf{Second, advantage calculation.} In
traditional agentic RL, each rollout is a Markov process
that maps to a unique training sample. In \add{\harl{}}, one rollout may
instead produce a dynamic number of training samples. This can result not only
from the retokenization issue described above, but also from harness operations
such as spawning subagents and summarizing context. These dynamic samples
challenge how rewards and advantages should be assigned to training samples.

\textbf{The third challenge is loss normalization.} In
\add{\harl{}}, a rollout may map to multiple training samples, making the
number of samples in each training batch dynamic. Loss normalization therefore
becomes nontrivial. For example, some existing frameworks still normalize
losses at the sample level, giving greater optimization weight to rollouts
that produce more samples, which may make the training unstable.

\textbf{The fourth challenge is training backend scheduling under dynamic sample counts.}
The number of samples produced by a rollout batch is known only after harness
execution and sample construction, while the number of training GPUs and their
parallel configuration remain fixed. The backend must partition this variable
sample set into training steps and mini-batches while balancing the workload
across fixed GPU workers.

In this work, we provide the first systematic characterization of these
challenges, and further present Agent Lightning v1.0, a complete refactoring of
the original Agent Lightning. It is a lightweight framework for RL training with arbitrary agent harnesses. Our design principle is to keep the system as simple as possible,
implemented in approximately 3,500 lines of code. Its training pipeline also
embeds our own design choices for the challenges described above, providing a
practical testbed for studying them.

We use Agent Lightning v1.0 to train general instruction-following agent, search agent, and coding
agent. In particular, existing agent frameworks provide limited
support for coding agent, including a lack of data and complete training
scripts, possibly because of the complexity of data cleaning, the difficulty
of environment setup, and the substantial computing resources required.
To address this gap,
we build on the open-source SWE-smith dataset and Qwen3.5-9B~\cite{qwen35} to provide a
complete data-cleaning pipeline and reproducible training scripts. Our final
RL run uses only 6K training samples and modest computing resources. Using RL
\add{alone, our trained model improves on SWE-bench Verified from 41.8\% to
56.4\%, an absolute 14.6\% gain.} We
release the complete workflow and scripts to the community to facilitate reproducibility.

\section{Challenges}
\label{sec:challenges}

Harnessed agentic RL changes how a rollout is observed and modeled by the
training engine. In traditional agentic RL, the training engine owns the
environment interaction loop and maintains the complete token history. Here,
$p_t$ denotes the prompt tokens at step $t$, $a_t$ denotes the action
corresponding to the response tokens, and $o_t$ is the tokenized environment
observation. The next prompt is constructed as
\begin{equation}
  p_t = (p_{t-1},a_{t-1},o_t).
\end{equation}
The overall rollout follows the sequence
$(p_1,a_1,o_1,a_2,o_2,a_3,\ldots)$. This forms a well-defined Markov process
and maps naturally to one linear training sample.

In \add{\harl{}}, the harness owns the environment interaction loop and
the message state. The training engine can only observe calls made through the
LLM endpoint. For a rollout $\rho$, it records a sequence
\begin{equation}
  \mathcal{C}(\rho)
  = \bigl((p_1,a_1),(p_2,a_2),\ldots,(p_{T_\rho},a_{T_\rho})\bigr),
\end{equation}
where $p_i$ is the prompt tokens and $a_i$ is the exact response tokens sampled by the model. The environment interactions and harness state
transitions between these calls are not directly visible. Consequently,
assembling the observed call sequence into training samples becomes a modeling
problem. This difference
introduces several implementation challenges. Existing frameworks make
different choices when addressing them, which can affect algorithmic
correctness and training stability.

\add{More formally, both traditional agentic RL and \harl{} admit a
partially observable Markov decision process formulation. Their distinction
lies in the latent state and the observation presented to the policy model. In
\harl{}, let
\begin{equation}
  s_t = \bigl(s_t^{\mathrm{harness}},s_t^{\mathrm{env}}\bigr)
\end{equation}
denote the latent execution state maintained jointly by the harness and the
environment. The model does not observe $s_t$ directly. Instead, the harness
constructs a message-level context and renders it into the exact token-level
prompt used for generation:
\begin{align}
  C_t^{\mathrm{msg}}
    &= \operatorname{Context}_{H}\bigl(s_t^{\mathrm{harness}}\bigr),\\
  p_t^{\mathrm{tok}}
    &= \operatorname{Tok}\bigl(\operatorname{Template}
       (C_t^{\mathrm{msg}})\bigr).
\end{align}
Each policy decision is therefore recorded as a call-level transition
\begin{equation}
  z_t = \bigl(p_t^{\mathrm{tok}},a_t^{\mathrm{tok}}\bigr),
  \qquad
  a_t^{\mathrm{tok}} \sim
  \pi_\theta\bigl(\cdot \mid p_t^{\mathrm{tok}}\bigr).
\end{equation}
A rollout yields a variable-length collection of such transitions, and no
exact token-prefix relation between consecutive prompts is assumed. Any
sequence construction performed for training must preserve the prompt under
which each recorded action was actually sampled. We next describe several new
challenges specific to \harl{}.}

\subsection{Retokenization and Sample Merging}

\paragraph{\add{Why Token-Prefix Continuity Breaks.}}

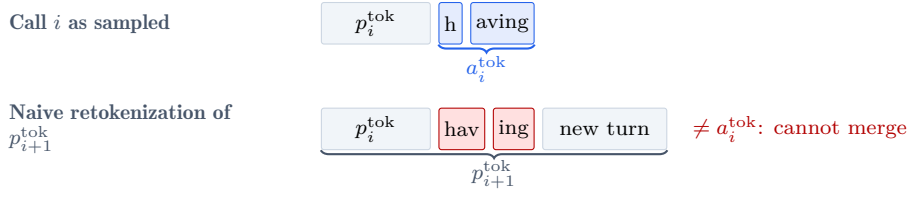
\begin{figure}[t]
  \centering
  \begin{tikzpicture}[
      tok/.style={draw, rounded corners=1pt, minimum height=0.55cm,
        inner sep=2pt, font=\scriptsize, align=center},
      ptok/.style={tok, fill=microsoftcard, draw=microsoftline,
        text width=1.3cm},
      newtok/.style={tok, fill=microsoftcard, draw=microsoftline,
        text width=1.5cm},
      atok/.style={tok, fill=microsoftblue!12, draw=microsoftblue},
      mismatch/.style={tok, fill=red!12, draw=red!70!black},
      rowlabel/.style={font=\scriptsize\bfseries, text=microsoftgray,
        align=left, text width=3.1cm},
    ]
    \node[rowlabel] (labelA) at (0,1.7) {Call $i$ as sampled};
    \node[ptok] (pA) at (3.3,1.7) {$p_i^{\mathrm{tok}}$};
    \node[atok, right=1mm of pA] (a1A) {h};
    \node[atok, right=1mm of a1A] (a2A) {aving};
    \draw[decorate, decoration={brace, amplitude=3pt, mirror},
      microsoftblue, thick]
      (a1A.south west) -- (a2A.south east)
      node[midway, yshift=-0.32cm, font=\scriptsize, text=microsoftblue]
      {$a_i^{\mathrm{tok}}$};

    \node[rowlabel] (labelB) at (0,0.3) {Naive retokenization of
      $p_{i+1}^{\mathrm{tok}}$};
    \node[ptok] (pB) at (3.3,0.3) {$p_i^{\mathrm{tok}}$};
    \node[mismatch, right=1mm of pB] (b1) {hav};
    \node[mismatch, right=1mm of b1] (b2) {ing};
    \node[newtok, right=1mm of b2] (bnew) {new turn};
    \draw[decorate, decoration={brace, amplitude=3pt, mirror},
      microsoftgray, thick]
      (pB.south west) -- (bnew.south east)
      node[midway, yshift=-0.32cm, font=\scriptsize, text=microsoftgray]
      {$p_{i+1}^{\mathrm{tok}}$};

    \node[font=\scriptsize, text=red!70!black, anchor=west]
      at ([xshift=2mm]bnew.east)
      {$\ne a_i^{\mathrm{tok}}$: cannot merge};
  \end{tikzpicture}
  \caption{A retokenization example. The word \texttt{having} sampled as the
  two tokens \texttt{h} and \texttt{aving} in call $i$ (\emph{top}) can be
  retokenized into different token boundaries, \texttt{hav} and \texttt{ing},
  when the updated message history is retokenized for call $i{+}1$
  (\emph{bottom}). Although the underlying text is identical, the token
  boundaries differ. This breaks the token-level prefix condition in
  Equation~\ref{eq:token-prefix} even though the text-level prefix condition
  in Equation~\ref{eq:text-prefix} still holds.}
  \label{fig:retokenization}
\end{figure}

Agent harness typically communicates with model APIs through text messages.
From the harness perspective, a rollout usually consists of turn-level calls
\begin{equation}
  \mathcal{C}^{\mathrm{text}}(\rho)
  = \bigl((p_1^{\mathrm{text}},a_1^{\mathrm{text}}),
  (p_2^{\mathrm{text}},a_2^{\mathrm{text}}),\ldots\bigr),
\end{equation}
where the harness sends $p_i^{\mathrm{text}}$ and receives
$a_i^{\mathrm{text}}$. Typically, for a multi-turn ReAct-style~\cite{react}
agent, the previous call forms a complete text-level prefix of the next
prompt, e.g.
\begin{equation}
  (p_i^{\mathrm{text}}, a_i^{\mathrm{text}}) \preceq p_{i+1}^{\mathrm{text}}.
  \label{eq:text-prefix}
\end{equation}
Besides, when the harness
spawns subagents or summarizes its context, the training engine may receive
subsequent calls with different prompt histories that do not satisfy this
relation.

RL training operates on exact token IDs and rollout log probabilities. The RL
system represents each call using the token sequences $p_i^{\mathrm{tok}}$ and
$a_i^{\mathrm{tok}}$, where $a_i^{\mathrm{tok}} \sim \pi_\theta(\cdot \mid
p_i^{\mathrm{tok}})$. The response returned to the harness is
$a_i^{\mathrm{text}}=\operatorname{Decode}(a_i^{\mathrm{tok}})$. Training
therefore observes the token-level sequence
$\mathcal{C}^{\mathrm{tok}}(\rho)=((p_1^{\mathrm{tok}},a_1^{\mathrm{tok}}),
(p_2^{\mathrm{tok}},a_2^{\mathrm{tok}}),\ldots)$.

\add{Token-prefix continuity between consecutive calls requires
\begin{equation}
  \bigl(p_i^{\mathrm{tok}},a_i^{\mathrm{tok}}\bigr)
  \preceq p_{i+1}^{\mathrm{tok}},
  \label{eq:token-prefix}
\end{equation}
where $\preceq$ denotes an exact token-level prefix.}

In practice, Equation~\ref{eq:text-prefix} holding does not guarantee that
Equation~\ref{eq:token-prefix} holds. The next request is
obtained by applying a chat template and tokenizer to the updated message
history. After retokenization, the token IDs corresponding to
$a_i^{\mathrm{text}}$ inside $p_{i+1}^{\mathrm{tok}}$ can differ from the
originally sampled IDs in $a_i^{\mathrm{tok}}$.

\add{This mismatch can arise through at least three mechanisms that we observe in our study:
\begin{enumerate}[leftmargin=*,label=(\arabic*)]
  \item \textbf{Chat-template non-compositionality.}
  Rendering a complete message history is not necessarily equivalent to
  concatenating the renderings of its parts:
  \begin{equation}
    \operatorname{Template}(A \mathbin{\Vert} B)
    \ne
    \operatorname{Template}(A)
    \mathbin{\Vert}
    \operatorname{Template}(B).
  \end{equation}
  A template may insert delimiters or connecting newlines at message
  boundaries, or omit markers that appeared in the original generation. In
  practice, for example, we find that Qwen's chat template can remove an
  earlier \texttt{<think>} marker, breaking token-prefix continuity.

  \item \textbf{Decode--retokenize drift.}
  Token decoding is not injective, so converting sampled tokens to text and
  tokenizing that text again need not recover the original token IDs:
  \begin{equation}
    \operatorname{Tok}
      \bigl(\operatorname{Decode}(a_i^{\mathrm{tok}})\bigr)
    \ne a_i^{\mathrm{tok}}.
  \end{equation}
  Figure~\ref{fig:retokenization} gives a concrete example: the word
  \texttt{having} is sampled as the two tokens \texttt{h} and
  \texttt{aving}, but retokenizing the same text inside a later prompt may
  instead produce \texttt{hav} and \texttt{ing}. Although the decoded text is
  unchanged, the token boundaries differ and break token-level continuity
  between the two calls.

  \item \textbf{Inference-time output transformation.}
  Tool-call and structured-output handlers may parse, normalize, repair, and
  reserialize a sampled response before returning it to the harness. Such
  processing can change whitespace, delimiters, JSON structure, or invalid
  syntax, so the response incorporated into a later prompt may differ even at
  the text level from the response represented by the sampled token IDs.
\end{enumerate}}

\paragraph{\add{Mitigating Broken Token-Prefix Continuity.}}

\add{The most direct strategy is to train every model call independently.
Each generated token is treated as part of its call's action, and the loss is
computed on $(p_i^{\mathrm{tok}},a_i^{\mathrm{tok}})$ without assuming any
relation to adjacent calls. This guarantees token-level correctness, but is
computationally inefficient because long prompt prefixes shared across calls
are repeatedly computed. Different frameworks handle it with different strategies.}

AReaL~\cite{areal2} and
verl Uni-Agent~\cite{uniagent} maintain a request buffer in the LLM proxy
that stores the historical text and tokens of each call. When a new request
arrives and its text matches the buffered history exactly, the framework
replaces the segment of the new prompt tokens that corresponds to the
previous response text with the buffered response tokens, which guarantees
that the token-prefix condition always holds. slime~\cite{slime} and
Polar~\cite{polar} do not perform this replacement.

\add{This replacement does more than recover computational reuse: when the
buffered token IDs differ from those in the actual next request, it changes
the prompt under which the next response is evaluated. Suppose the actual
next prompt contains a reconstructed version
$\widehat{a}_i^{\mathrm{tok}}$ of the previous response:
\begin{equation}
  p_{i+1}^{\mathrm{tok}}
  =
  p_i^{\mathrm{tok}}
  \mathbin{\Vert}
  \widehat{a}_i^{\mathrm{tok}}
  \mathbin{\Vert}
  \Delta_{i+1}.
\end{equation}
Replacing this segment with the originally sampled
$a_i^{\mathrm{tok}}$ produces a stitched prompt
\begin{equation}
  \widetilde{p}_{i+1}^{\mathrm{tok}}
  =
  p_i^{\mathrm{tok}}
  \mathbin{\Vert}
  a_i^{\mathrm{tok}}
  \mathbin{\Vert}
  \Delta_{i+1},
  \qquad
  \widetilde{p}_{i+1}^{\mathrm{tok}}
  \ne p_{i+1}^{\mathrm{tok}}.
\end{equation}
The response $a_{i+1}^{\mathrm{tok}}$ was sampled from the policy conditioned
on $p_{i+1}^{\mathrm{tok}}$, not on
$\widetilde{p}_{i+1}^{\mathrm{tok}}$. Training it under the stitched prompt
therefore introduces an off-policy discrepancy. Exact token-prefix overlap
should be used for merging only when it preserves the prompt actually
consumed during rollout.}

\add{A second approach is prefix-shared or tree-structured
training. Exact common token prefixes are represented once, while a
branch-aware causal attention mask ensures that tokens attend only to their
ancestors and earlier tokens on the same branch. This can reproduce the
result of training independent causal sequences while reusing prefix
computation. However, tree packing, custom attention masks or kernels,
partitioning, and distributed gradient handling require substantial training
backend support.}

\add{A third practical approach is best-effort sequence merging. Two
consecutive calls are merged only when their observed token IDs satisfy
Equation~\ref{eq:token-prefix}. When the condition holds, only the unmatched
suffix and next action are appended. When it fails, the current sequence is
closed and a new one begins. This preserves the prompts consumed during
rollout and works with standard dense causal kernels, while retokenization
drift merely lowers the merge ratio. Agent Lightning v1.0 adopts this strategy as a middle
ground between independent-call recomputation and backend-intensive tree
training.}

\add{These approaches have different trade-offs. Buffered token
replacement can increase the merge ratio, but becomes off-policy stitching
when it changes the prompt actually consumed during rollout. Independent-call
training and best-effort merging guarantee token-level correctness but may leave
redundant prefix computation, whereas tree-structured training can recover
more reuse at the cost of a substantially more complex backend.}

\subsection{Advantage Calculation}
\label{subsec:advantage-calculation}

After LLM calls are merged, one rollout $\rho$ can produce a
different number of training samples $N_\rho$, known only after execution and
sample construction. Retokenization is one source of
this dynamic sample count. The harness can
also spawn subagents, creating branches that do not share one linear history,
or summarize its context, replacing the previous token prefix with a new one.
These operations can split one task-level rollout into multiple training
samples. This is not an edge case in practice: in our coding-agent training
runs (Figure~\ref{fig:smith-merge-sample-metrics}), only 36\% of rollouts on
average remain as a single training sample, and each rollout yields 2.4
training samples on average.

In practice, reward is still outcome-based and is assigned to every sample
within the rollout that produced it. This raises a natural question: when
computing advantage, should the group statistics be computed at the rollout
level or the sample level? We find that existing frameworks make different
choices. verl Uni-Agent~\cite{uniagent} and Polar~\cite{polar} compute
advantage at the rollout level, while slime~\cite{slime} and
AReaL~\cite{areal2} compute it at the sample level.

\begin{figure}[t]
  \centering
  \includegraphics[width=\linewidth]{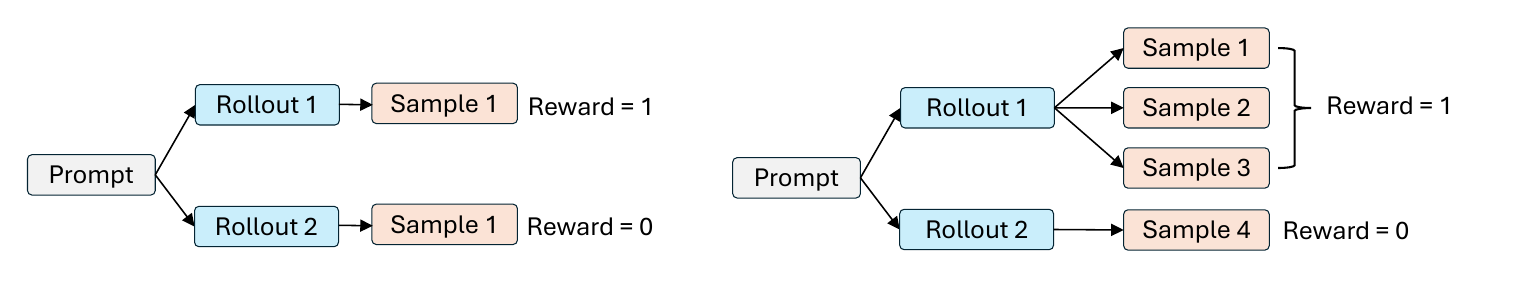}
  \caption{Traditional agentic RL, where each rollout is one training sample
  (\emph{left}), versus \add{\harl{}}, where a rollout can expand into a
  dynamic number of samples that inherit its reward (\emph{right}).}
  \label{fig:rollout-sample}
\end{figure}

Figure~\ref{fig:rollout-sample} gives a concrete example. Suppose Rollout 1
and Rollout 2 are generated from the same prompt in one
GRPO~\cite{deepseekmath} group, with Rollout 1 receiving reward 1 and Rollout
2 receiving reward 0. In traditional RL (\emph{left}), each rollout maps to
exactly one training sample, and computing the baseline as
$\bar{r}=(1+0)/2=1/2$ is uncontroversial. In \add{\harl{}}
(\emph{right}), Rollout 1 has three samples (Sample 1, Sample 2, and Sample
3) while Rollout 2 remains a single sample (Sample 4): rollout-level
advantage calculation still gives baseline
$\bar{r}_{\mathrm{rollout}}=(1+0)/2=1/2$, whereas sample-level advantage
calculation instead gives $\bar{r}_{\mathrm{sample}}=(1+1+1+0)/4=3/4$.

We believe rollout-level advantage is the more principled choice, for the
following reasons. Retokenization is an incidental phenomenon, and advantage
assignment should
not change simply because retokenization happened to split a rollout into
more samples. Likewise, subagent spawning and context summarization are
internal operations of the harness and should not be allowed to change the
baseline of the entire group. Future work may still be needed to design
better credit assignment across the samples within a rollout.

\subsection{Loss Normalization}
\label{subsec:loss-normalization}

Dynamic sample counts also make loss normalization a nontrivial choice.
Consider a training batch of $R$ rollouts, where rollout $\rho$ produces
$N_\rho$ samples, and sample $j$ of rollout $\rho$ has $L_{\rho,j}$ response
tokens with per-token loss $\ell_{\rho,j,t}$ for $t=1,\ldots,L_{\rho,j}$. Most
frameworks directly reuse a sample-level normalization inherited from
traditional agentic RL. The first is the
token-mean loss used by DAPO~\cite{dapo}, which sums the loss over every
token in the batch and normalizes by the total number of response tokens:
\begin{equation}
  \mathcal{L}_{\mathrm{token\text{-}mean}}
  = \frac{\sum_{\rho=1}^{R}\sum_{j=1}^{N_\rho}\sum_{t=1}^{L_{\rho,j}}
    \ell_{\rho,j,t}}
    {\sum_{\rho=1}^{R}\sum_{j=1}^{N_\rho} L_{\rho,j}}.
  \label{eq:token-mean}
\end{equation}
The second is the seq-mean-token-mean loss used by GRPO~\cite{deepseekmath},
which first averages the loss within each sample and then averages these
sample means uniformly over all samples in the batch:
\begin{equation}
  \mathcal{L}_{\mathrm{seq\text{-}mean}}
  = \frac{1}{\sum_{\rho=1}^{R} N_\rho}
    \sum_{\rho=1}^{R}\sum_{j=1}^{N_\rho}
    \frac{1}{L_{\rho,j}}\sum_{t=1}^{L_{\rho,j}} \ell_{\rho,j,t}.
  \label{eq:seq-mean}
\end{equation}
slime~\cite{slime} implements a rollout-level token-mean loss, which first pools all
response tokens of a rollout together and then averages uniformly over
rollouts:
\begin{equation}
  \mathcal{L}_{\mathrm{rollout\text{-}mean}}
  = \frac{1}{R}\sum_{\rho=1}^{R}
    \frac{\sum_{j=1}^{N_\rho}\sum_{t=1}^{L_{\rho,j}} \ell_{\rho,j,t}}
    {\sum_{j=1}^{N_\rho} L_{\rho,j}}.
  \label{eq:rollout-mean}
\end{equation}

\begin{figure}[t]
  \centering
  \includegraphics[width=0.65\linewidth]{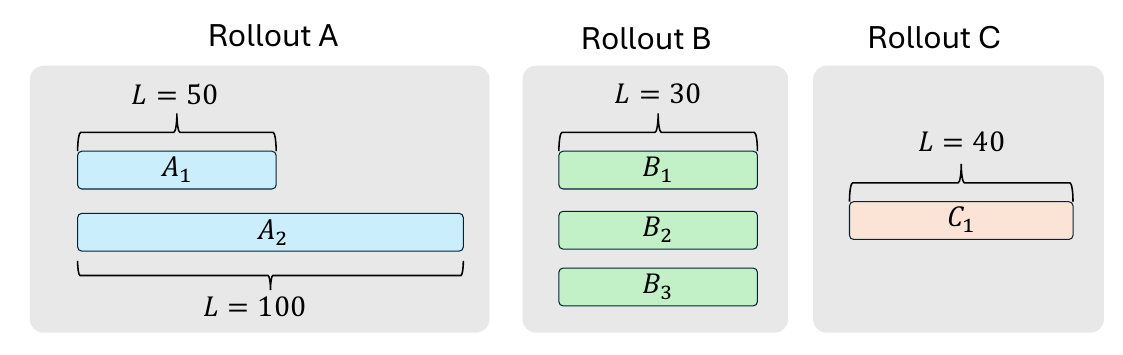}
  \caption{An example batch with three rollouts of different sample counts
  and response lengths.}
  \label{fig:loss-normalization}
\end{figure}

We provide a more concrete example in Figure~\ref{fig:loss-normalization}.
Suppose a batch contains three rollouts: Rollout A produces two samples
$A_1$ and $A_2$ with response lengths 50 and 100; Rollout B produces three
samples $B_1$, $B_2$, and $B_3$, each with response length 30; Rollout C
produces a single sample $C_1$ with response length 40. Let
$A_1,A_2,B_1,B_2,B_3,C_1$ also denote the sum of per-token losses within
each sample, i.e.\ $A_1=\sum_t \ell_{A,1,t}$, and so on. Then:
\begin{itemize}
  \item $\mathcal{L}_{\mathrm{token\text{-}mean}}
  = (A_1+A_2+B_1+B_2+B_3+C_1)/(50+100+30+30+30+40)$.
  \item $\mathcal{L}_{\mathrm{seq\text{-}mean}}
  = \tfrac{1}{6}(A_1/50+A_2/100+B_1/30+B_2/30+B_3/30+C_1/40)$.
  \item $\mathcal{L}_{\mathrm{rollout\text{-}mean}}
  = \tfrac{1}{3}\bigl((A_1+A_2)/(50+100)+(B_1+B_2+B_3)/(30+30+30)+C_1/40\bigr)$.
\end{itemize}

For loss normalization, we hold the same view as for advantage calculation:
sample count should not be allowed to affect gradient normalization, because
it is often driven by incidental factors such as retokenization. Under this
view, the
seq-mean-token-mean loss in Equation~\ref{eq:seq-mean} is problematic because
it varies with how many samples a single rollout happens to produce, and in
general gives disproportionately more weight to rollouts with more samples.
We therefore believe the token-mean loss in Equation~\ref{eq:token-mean} and
the rollout-level token-mean loss in Equation~\ref{eq:rollout-mean} are more
principled in theory. In
practice, however, we find that the token-mean loss is sensitive to long
sequences: when many long negative samples appear in a batch, it can cause
instability later in training. We therefore prefer the rollout-level
token-mean loss in Equation~\ref{eq:rollout-mean}.

\subsection{Training Backend Complexity}

Dynamic sample counts also complicate the interface with the training backend.
The number and lengths of samples in a rollout batch are known only after
harness execution and sample construction. In contrast, the number of training
GPUs and the data-, tensor-, and pipeline-parallel configuration are typically
fixed throughout training. The backend must therefore map a variable workload
onto a fixed set of workers at every iteration.

\add{After sample construction, the backend may flatten the resulting
sequences into a physical tensor batch, but this transformation must preserve
their statistical provenance. In particular, every sequence should retain its
rollout identifier and prompt-group identifier:
\begin{equation}
  \mathcal{B}_{\mathrm{train}}
  =
  \bigcup_{\rho \in \mathcal{B}_{\mathrm{rollout}}}
  \left\{
    \bigl(S_{\rho,j},\rho,g_\rho\bigr)
    \;\middle|\;
    1 \le j \le N_\rho
  \right\},
\end{equation}
where $S_{\rho,j}$ is the $j$-th sequence constructed from rollout $\rho$ and
$g_\rho$ identifies the group of rollouts sampled from the same prompt.
Flattening changes only the physical representation; it must not change
rollout membership or cause a rollout to receive additional statistical
weight merely because it produced more sequences.}

\add{Rollout boundaries also constrain batch scheduling. Row-based tensor
batches, data-parallel partitions, and micro-batch schedules cannot be planned
from the prompt or rollout count alone because $N_\rho$ is known only after
execution. Moreover, sequences from one rollout should remain in the same
optimizer update. Splitting them across updates would evaluate different
parts of one rollout under different policy versions, introducing
within-rollout policy skew. A backend must therefore balance token workload
across fixed workers while preserving these rollout-level statistical and
update boundaries.}

\section{System Design}
\label{sec:system-design}

\add{When the trainer and agent harness are disaggregated, no single process
owns the complete rollout lifecycle. The trainer owns model inference and
optimization, while the harness owns context construction, control flow, tool
use, and environment interaction. Agent execution may run remotely, persist
beyond any single API request or worker process, and fail independently of
the training process.
Operationalizing this architecture therefore requires a lightweight control
plane that coordinates durable rollout state, external execution, partial
failures, and resource usage without pulling harness logic back into the
trainer.}

\add{Agent Lightning v1.0 builds this control plane around a declarative rollout
abstraction and a reconciliation loop. The trainer declares rollouts through
the API Gateway, which serves as the source of truth for lifecycle state and
append-only events. The Rollout Controller continuously reconciles this state
with agent executions running as Kubernetes Jobs or local processes. This
separation makes Kubernetes an interchangeable execution backend rather than
part of the rollout abstraction itself.}

\add{The same control plane provides explicit reliability and observability
semantics across the service boundary. Control-plane operations are
idempotent, generation attempts are recorded and resolved explicitly, and a
rollout identifier links model requests, rewards, custom events, and execution
logs into one diagnostic record. Finally, the API Gateway coordinates
inference admission during collocated asynchronous RL, allowing rollout and
weight update to time-share one GPU pool without exposing phase switches to
the external harness.}

As shown in Figure~\ref{fig:teaser}, Agent Lightning v1.0 bridges the training cluster (a
GPU cluster running model inference and training) and the agent execution
cluster through three components. The API Gateway is an API service that
stores rollouts, models, and events, and forwards LLM calls from agent
harnesses to the model endpoints the trainer has registered. The Rollout
Controller manages agent execution on top of a Kubernetes cluster (or a
local process pool), polling rollouts from the API Gateway and launching the
corresponding agent tasks. The Customized Trainer, built on top of
VERL~\cite{hybridflow}, registers rollouts with the API Gateway, waits for
the Rollout Controller to drive them to completion, and then retrieves their
recorded events to assemble training samples. Through this chain, the trainer
only creates rollouts and collects trajectories, any harness can connect by
switching its LLM endpoint to the proxy, and training and execution resources
can be provisioned independently and even run in different locations.

Agent Lightning v1.0 is designed to be as simple as possible: the whole system is
implemented in approximately 3,500 lines of code, with each component having a
clear responsibility. Agent Lightning v1.0 also incorporates our own
design choices for the challenges discussed in
Section~\ref{sec:challenges}, such as rollout-level reward and advantage
calculation and rollout-level loss normalization.

We describe each component in detail in
Appendix~\ref{sec:appendix-system-design}, and highlight several features in the following sections.

\subsection{Collocated Async RL}

In agentic RL, the synchronous RL setup is slow:
all rollouts in a batch must finish before the training step can update the
model, so the training step must wait for the slowest rollout in the batch, leaving many
GPUs idle. Asynchronous RL,
proposed by AReaL~\cite{areal}, solves
the GPU idling problem by splitting rollout and update onto two separate
pools of machines, each occupying different GPUs, so the rollout GPUs can
keep working while the update is running. However, we find that this requires more GPUs overall, which demands more
resources than smaller teams may be able to afford. It also requires
separately managing a rollout queue and an update queue, which is more
complex because the two queues can progress at different rates in
practice.

\begin{figure}[t]
	\centering
	\includegraphics[width=\linewidth]{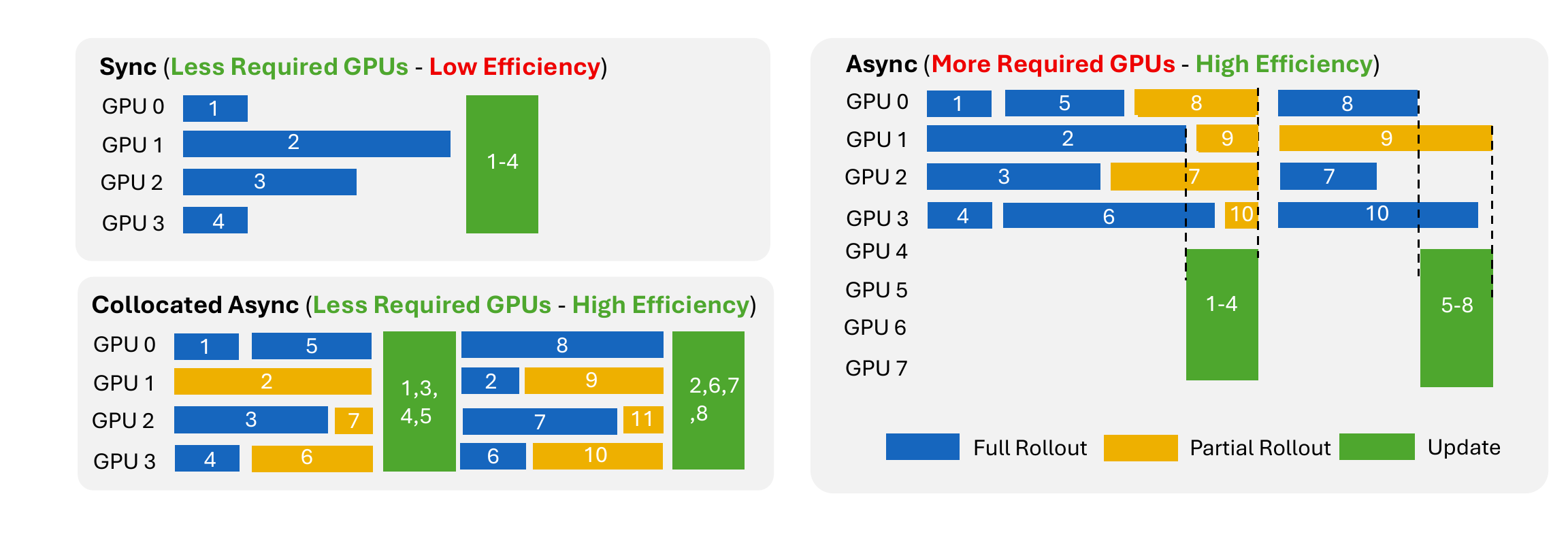}
	\caption{Sync RL, async RL, and our collocated async RL. Collocated
	async RL shares the same GPUs between rollout and update while still
	avoiding the need to wait for the slowest rollout.}
	\label{fig:collocated-async}
\end{figure}

We instead propose \emph{collocated async}, illustrated in
Figure~\ref{fig:collocated-async}. In collocated async RL, rollout and weight
update share the same pool of GPUs. Once enough rollout data has been collected, the update step begins.
The API Gateway simultaneously stops accepting new requests and waits for the current ones
to complete. If a new request arrives afterward, the Gateway pauses it
until the system re-enters the rollout phase. As a result, the switch is
invisible to the agent harness.

Collocated async lets the same GPUs be time-shared between rollout and
training, so we use fewer GPUs than async RL, while also avoiding the wait
for the slowest rollout. In our experiments, collocated async RL achieves
roughly a 2x end-to-end speedup over synchronous RL while also using fewer
GPUs.

\subsection{Network Issues}

In \add{\harl{}}, agents are separate from the training engine, which
causes network issues: calls between the two sides travel over the
network and are not always reliable. Two kinds of calls are affected: calls
from the Trainer or the Rollout Controller to the API Gateway, and
calls from the agent harness to the LLM inference endpoint through the
proxy. Both can experience network interruptions, and the calling side
commonly retries after such a failure. We address this with two measures.

\paragraph{Idempotent API Gateway endpoints.} We design every rollout API
Gateway endpoint to be idempotent, so that repeating the same call any
number of times has the same effect as calling it once. This lets a caller
retry freely after a network failure without worrying that the retry
itself will corrupt state.

\paragraph{Deduplication of repeated LLM API calls.} A retried LLM API call
cannot be made idempotent in the same way, since each retry is a new
generation request and may return a different response. Instead, when the
Customized Trainer assembles training samples, it deduplicates
\texttt{model\_request} events that share the same prompt: if a rollout
recorded multiple calls with an identical prompt, only the last (most
recent) call is kept, and the earlier ones, which correspond to retried or
superseded calls, are discarded.

\subsection{Kubernetes Integration}

During the rollout phase of RL training, many agents must run
concurrently, which requires substantial compute resources.
Existing \add{\harl{}} frameworks commonly turn to commercial sandbox
services to meet this demand. For example, verl Uni-Agent~\cite{uniagent}
launches agent execution on managed sandbox offerings such as Modal
Sandbox and Volcano veFaas, while slime~\cite{slime} relies on
E2B. These services provide convenient, ready-to-use sandboxing, but they can be
expensive at the scale of RL training.

Agent Lightning v1.0 instead runs agents directly on a Kubernetes cluster
through its Rollout Controller. Each agent execution is scheduled as a
standard Kubernetes Job, so users can rely entirely on self-hosted or
on-premise compute rather than a commercial sandbox provider. This avoids
the recurring cost of commercial sandbox services and keeps the entire
training stack open-source.

\subsection{Monitoring}

During training, we find that agents themselves can run into problems such
as reward hacking, bad agent behavior, and network connectivity issues.
Manually inspecting rollouts to catch these problems is inconvenient, so
the Customized Trainer includes a monitoring system that records training
and validation rollouts, together with pod-level logs from Kubernetes,
described further in Appendix~\ref{sec:appendix-system-design}. This
system lets us use AI agents to automatically identify such issues, and we
have indeed found several reward-hacking examples this way, as shown in
Section~\ref{subsec:reward-hacking}.

\section{Experiments}

We evaluate Agent Lightning v1.0 in three practical agent training settings: search,
general instruction following, and coding. We follow the experimental setup of
Search-R1~\cite{searchr1} to train search agents and the setup of
LLM-in-Sandbox~\cite{llminsandbox} to train general instruction-following
agents. For coding agents, we build our training data from
SWE-smith~\cite{swesmith}.

We find that existing frameworks rarely provide a complete, reproducible
coding-agent training example, possibly because of the complexity of data
cleaning, the difficulty of environment setup, and the substantial computing
resources required. We therefore focus on the coding agent and describe its
training process in detail, aiming to provide a complete and reproducible
example that requires only modest resources.

\subsection{Search Agent}

We follow the experimental setting of Search-R1~\cite{searchr1} to train a
search agent that interleaves reasoning with search-engine queries and uses
retrieved passages to answer knowledge-intensive questions. We use
Llama-3.2-3B-Instruct~\cite{llama3} as the policy model and optimize it with
GRPO~\cite{deepseekmath}. We train on the training split of
HotpotQA~\cite{hotpotqa}. For evaluation, we sample 50 examples from each of
HotpotQA, 2WikiMultiHopQA~\cite{2wikimultihopqa},
MuSiQue~\cite{musique}, Bamboogle~\cite{bamboogle},
TriviaQA~\cite{triviaqa}, and Natural Questions~\cite{naturalquestions}. We set
the training batch size to 512, sample 4 rollouts per prompt, and evaluate
the model every 10 training steps. We use exact match (EM) as the reward metric.

\begin{figure}[t]
  \centering
  \includegraphics[width=0.65\textwidth]{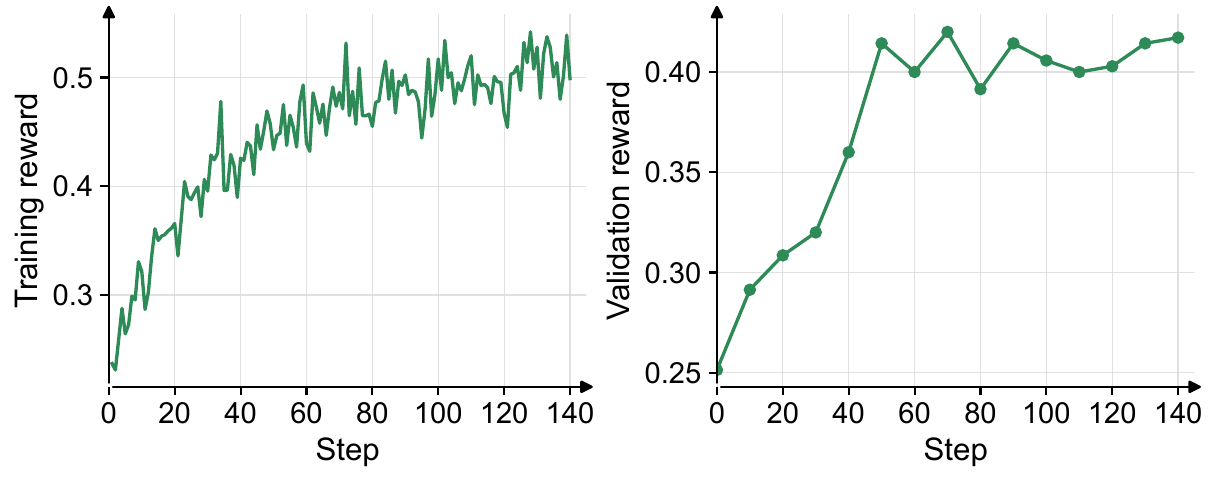}
  \caption{Search-agent training dynamics. From left to
  right: mean training reward and mean validation reward.}
  \label{fig:search-r1-training}
\end{figure}

Figure~\ref{fig:search-r1-training} summarizes the training dynamics. The mean
training reward increases steadily. On the validation set, the reward improves
from 25.1\% to 41.7\%, an absolute 16.6\% gain.

\subsection{General Instruction-Following Agent}

We follow the experimental setting of LLM-in-Sandbox~\cite{llminsandbox} to
train a general instruction-following agent. The agent solves diverse
non-coding tasks by using a computer sandbox to access external resources,
manage files, and execute code. We use the agent harness provided by the
original authors. We use
Qwen3-4B-Instruct-2507~\cite{qwen3} as the policy model and optimize it with
RLOO~\cite{rloo}. We use the dataset released by Instruction
Pre-Training~\cite{instructionpretraining} and split it into 80\% for training
and 20\% for evaluation. We set the training batch size to 8, sample 8 rollouts
per prompt, and evaluate the model every 20 training steps.

\begin{figure}[t]
  \centering
  \includegraphics[width=0.65\textwidth]{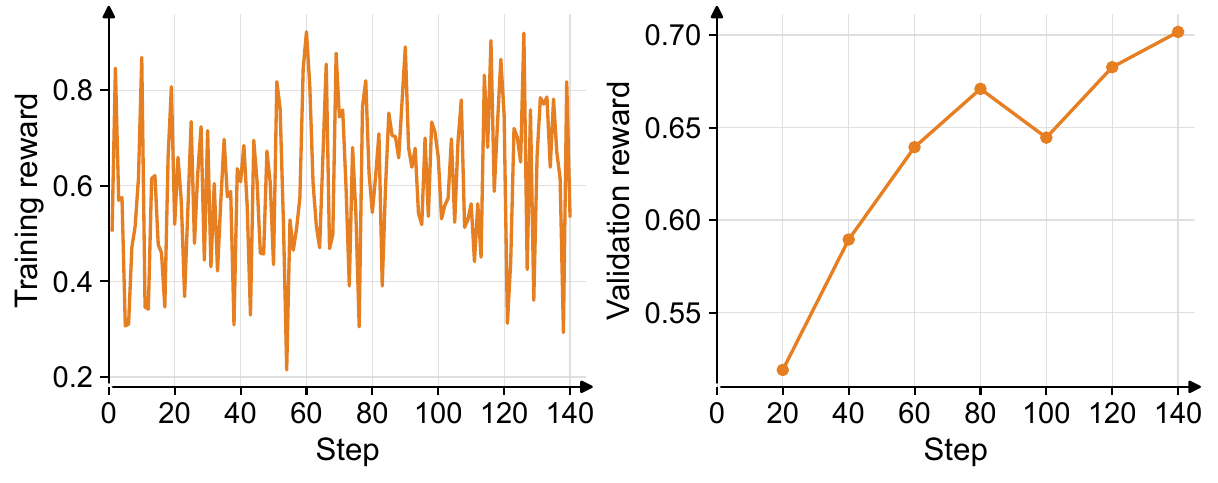}
  \caption{General instruction-following agent training dynamics. From left to
  right: mean training reward and mean validation reward.}
  \label{fig:llm-in-sandbox-training}
\end{figure}

Figure~\ref{fig:llm-in-sandbox-training} shows that the batch-level training
reward is noisy, whereas the validation reward exhibits a clear upward trend.
It improves from 51.9\% to 70.2\%, an absolute 18.3\% improvement.

\subsection{Coding Agent}

We train a coding agent based on Qwen3.5-9B~\cite{qwen35} using tasks derived from the
SWE-smith dataset~\cite{swesmith}. We use mini-SWE-agent~\cite{minisweagent}
as the agent harness for interacting with repository environments, executing
commands, and producing code changes. To obtain reliable training signals, we
apply a detailed data-filtering pipeline, introduce safeguards against reward
hacking, and implement the additional measures required for robust coding-agent
training. The following sections describe these implementation details.

\subsubsection{Dataset Preprocessing and Filtering}

SWE-smith is a large-scale dataset of executable software-engineering tasks
constructed by introducing bugs into real Python repositories. It contains
59,136 tasks from 128 repositories, with problem statements, code patches, and
tests for verifying solutions~\cite{swesmith}. Its Docker images occupy only
295~GB ~\cite{swesmith}, substantially less than the
4~TB required by R2E-Gym~\cite{r2egym} and the 6~TB required by
SWE-Gym~\cite{swegym}.

For each task, SWE-smith first switches the repository to the corresponding
problem branch and asks the coding agent to modify the codebase. It then runs a
task-specific test suite to determine whether the submitted changes resolve the
problem.

We identify the following issues in the released data:
\begin{itemize}
  \item Among the 59,136 records, 18,033 have an empty problem statement.
  \item For 1,265 records, the corresponding problem branch is missing from
  the provided Docker image.
  \item Some tasks require large test suites. For example,
  \texttt{python-jsonschema} requires executing more than 7,000 tests, consuming
  substantial CPU and memory.
\end{itemize}

We therefore remove tasks with an empty problem statement, a missing problem
branch, or more than 200 tests. The remaining tasks still have a highly skewed difficulty distribution and provide
limited training signal, so we apply an additional model-based difficulty
filter. We run Qwen3.5-9B four times on every candidate: tasks solved in all four
rollouts are removed, while tasks with both successful and failed rollouts are
retained, yielding approximately 5,000 examples. To avoid making the resulting
set overly easy, we additionally sample 1,000 tasks that fail in all four
rollouts. The final split contains approximately 6,000 training examples and
400 test examples.

\subsubsection{Preventing Reward Hacking}
\label{subsec:reward-hacking}

During training, we observe several reward-hacking behaviors in which the agent
bypasses the intended problem-solving process and obtains the reference source
code directly:
\begin{enumerate}
  \item Using Git history to locate the gold commit.
  \item Using \texttt{wget} or \texttt{curl} to retrieve upstream source code
  from GitHub.
  \item Using \texttt{pip} to download a package's source code.
  \item Using Python networking libraries, such as \texttt{urllib}, to download
  source code.
\end{enumerate}

We introduce two safeguards. First, we disable Git commands and hide the
\texttt{.git} directory from the agent, preventing it from inspecting commit
history. Second, we enforce a Kubernetes network policy that blocks general
outbound network access and permits connections only to explicitly whitelisted
services. Together, these measures require the agent to solve each task using
only the provided problem statement and local information.

\subsubsection{Training Dynamics}

\begin{figure}[t]
  \centering
  \includegraphics[width=0.75\textwidth]{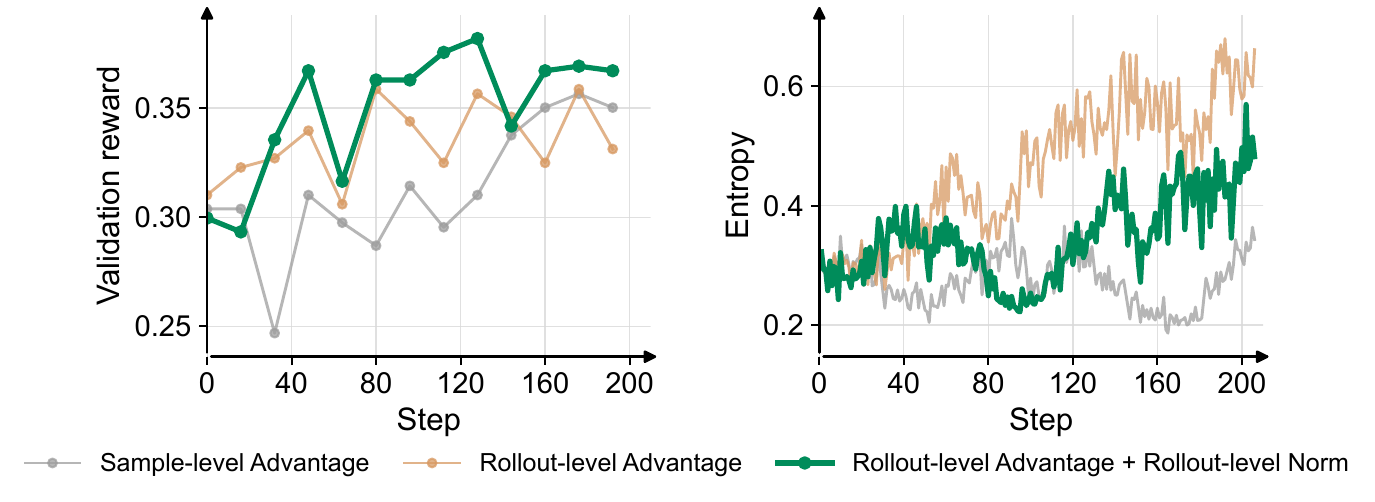}
  \caption{Coding-agent training dynamics for Sample-level Advantage,
  Rollout-level Advantage, and Rollout-level Advantage + Rollout-level Norm.
  Left: validation reward. Right: policy entropy.}
  \label{fig:smith-ablation-training}
\end{figure}

\begin{figure}[t]
  \centering
  \includegraphics[width=0.75\textwidth]{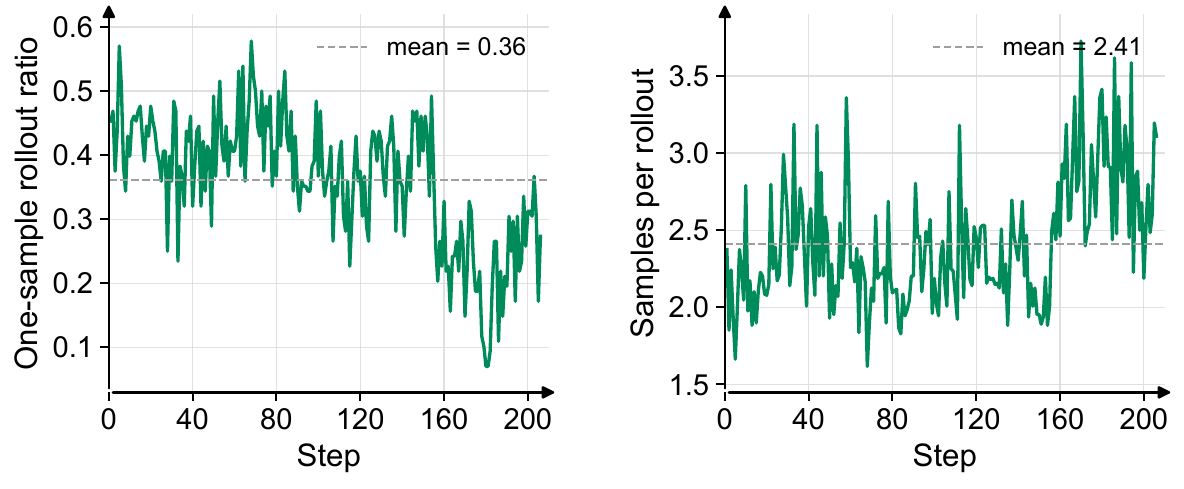}
  \caption{Rollout-merging behavior for the Rollout-level Advantage +
  Rollout-level Norm run. Left: fraction of rollouts that yield exactly one
  training sample. Right: average number of training samples produced per
  rollout. Dashed lines mark the mean over training.}
  \label{fig:smith-merge-sample-metrics}
\end{figure}

As discussed in Section~\ref{sec:challenges}, because a rollout's sample count is
often driven by incidental factors such as retokenization, both the advantage
calculation (Section~\ref{subsec:advantage-calculation}) and the loss
normalization (Section~\ref{subsec:loss-normalization}) should be computed at
the rollout level rather than the sample level. We validate this design
choice on our coding-agent training run by comparing three settings, all
using the same underlying GRPO objective:
\begin{itemize}
  \item \emph{Sample-level Advantage}, which combines sample-level advantage
  with the token-mean loss in Equation~\ref{eq:token-mean};
  \item \emph{Rollout-level Advantage}, which switches only the advantage
  calculation to the rollout level while keeping the token-mean loss; and
  \item \emph{Rollout-level Advantage + Rollout-level Norm}, which
  additionally replaces the token-mean loss with the rollout-level
  token-mean loss in Equation~\ref{eq:rollout-mean}.
\end{itemize}
Figure~\ref{fig:smith-ablation-training}
compares the three settings. The last variant
produces the highest observed validation reward, reaching 38.2\% at step 128,
compared with 35.0\% for the baseline and 33.1\% when only the
rollout-advantage fix is applied. Its policy entropy also grows more slowly
and stays more stable over training than the variant with only the
rollout-advantage fix. These results suggest that loss normalization
controls the entropy increase introduced by the corrected rollout advantages
while improving validation reward. We also evaluate the Rollout-level
Advantage + Rollout-level Norm checkpoint on SWE-bench
Verified~\cite{swebench}, where it improves from 41.8\% to 56.4\% at step
208. Because coding-agent trajectories vary widely in length, the
rollout-level advantage variants merge trajectories into training rows
wherever possible to reduce padding waste, which produces a dynamic number
of training samples per rollout: on average, only 36\% of rollouts remain as
a single, fully merged row, and each rollout yields 2.41 training samples on
average (Figure~\ref{fig:smith-merge-sample-metrics}).

\section{Related Work}

Traditional RL frameworks such as verl~\cite{hybridflow}, AReaL~\cite{areal},
and slime~\cite{slime} originally required the agent loop to be implemented
directly inside the training framework, following the classic
ReAct-style~\cite{react} Markov formulation in which the training engine owns
the environment interaction loop. This makes it difficult to reuse
independently maintained agent harnesses, such as
mini-SWE-agent~\cite{minisweagent}, OpenHands~\cite{openhands},
OpenCode~\cite{opencode}, Claude Code~\cite{claudecode}, Codex~\cite{codex},
OpenClaw~\cite{openclaw}, and Hermes~\cite{hermes}, since each would need to be
reimplemented inside the training stack. Our original Agent
Lightning~\cite{agentlightning} work introduced a disaggregated architecture
that instead connects arbitrary agent harnesses to RL training through an LLM
endpoint, and this proxy-based approach has since been adopted by verl
Uni-Agent~\cite{uniagent}, AReaL 2.0~\cite{areal2}, slime
v0.3.0~\cite{slime}, and Polar~\cite{polar}. As discussed in
Section~\ref{sec:challenges}, these frameworks make different, sometimes
conflicting, design choices when handling retokenization, advantage
calculation, and loss normalization under a dynamic number of training
samples per rollout, and they commonly rely on commercial sandbox services,
such as Modal Sandbox, Volcano veFaas, and E2B, to execute agents at scale.
Agent Lightning v1.0 instead runs entirely on a self-hosted Kubernetes cluster and
implements the whole system in approximately 3,500 lines of code, providing a compact
and transparent testbed for studying these design choices, which we validate
on the search-agent, instruction-following, and coding-agent settings of
Search-R1~\cite{searchr1}, LLM-in-Sandbox~\cite{llminsandbox}, and
SWE-smith~\cite{swesmith}, respectively.

\section{Conclusion}

We characterize \emph{\add{\harl{}}}, a paradigm in which the
deploy-time agent harness, not the training engine, owns the environment
interaction loop, and identify the resulting challenges in retokenization,
advantage calculation, loss normalization, and training backend scheduling.
We present Agent Lightning v1.0, an approximately 3,500-line framework that supports arbitrary agent
harnesses and embeds our rollout-level design choices for these challenges.
We validate it on search, instruction-following, and coding agents, and
release a complete data pipeline and reward-hacking safeguards that let RL
\add{alone improve Qwen3.5-9B on SWE-bench Verified from 41.8\% to 56.4\%, a
gain of 14.6 percentage points, using only about 6K training examples.} We
release the full codebase and scripts to facilitate
reproducible \add{\harl{}} research.

\clearpage
\bibliographystyle{unsrtnat}
\bibliography{reference}

\clearpage
\appendix
\section{Detailed System Design}
\label{sec:appendix-system-design}

This appendix provides a detailed description of Agent Lightning v1.0's three
components introduced in Section~\ref{sec:system-design}: the API Gateway,
the Rollout Controller, and the Customized Trainer.

\subsection{API Gateway}

\begin{figure}[t]
	\centering
	\includegraphics[width=0.5 \linewidth]{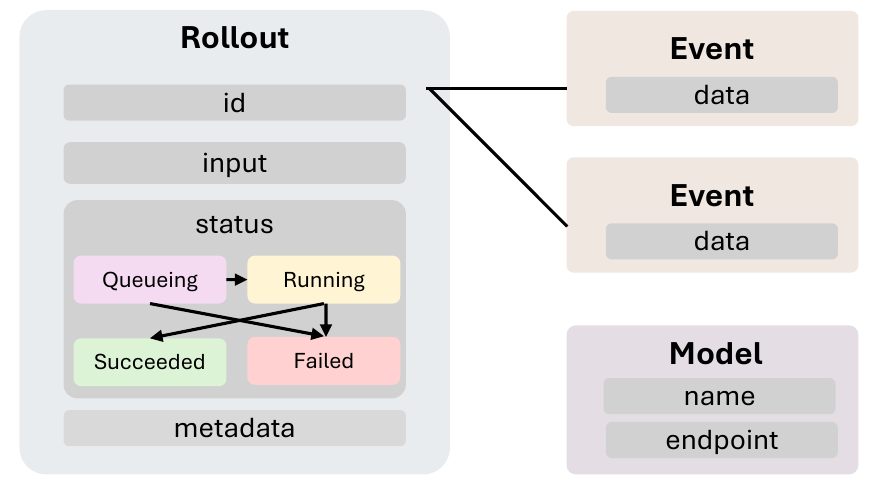}
	\caption{The objects stored by the API Gateway. }
	\label{fig:api-gateway-schema}
\end{figure}

The API Gateway is the central component of Agent Lightning v1.0, kept deliberately
lightweight: a single stateful service that stores rollouts, models, and
events, and exposes them through a minimal API. Figure~\ref{fig:api-gateway-schema}
summarizes these objects and their relationships.

\paragraph{Rollout.}
A \emph{rollout} is one agent execution, identified by a unique rollout ID. It
stores an input (derived from a training example), a status, and
user-defined metadata. The status follows the state machine in
Figure~\ref{fig:api-gateway-schema}: \texttt{queuing}, \texttt{running}, and
the terminal states \texttt{succeeded} and \texttt{failed}. Rollouts are not
one-to-one with training examples: GRPO~\cite{deepseekmath}, for instance,
generates multiple independent rollouts, each with its own ID and trajectory,
from the same example.

\paragraph{Model.}
A \emph{model} identifies an LLM inference endpoint by its name and address.
The trainer registers models with the API Gateway, which then routes agent
harness requests to the corresponding inference server.

\paragraph{Event.}
An \emph{event} attaches arbitrary data to a rollout. By default, Agent Lightning v1.0
records a \texttt{model\_request} event (prompt token IDs, response token IDs,
and response log probabilities) for every LLM interaction, and a
\texttt{reward} event that the agent typically reports once at the end of the
rollout with a scalar reward. Users can also define custom event types.

\medskip

Table~\ref{tab:api-gateway-endpoints} lists the API Gateway's endpoints,
which fall into two APIs: the rollout API and the proxy API.

\begin{table}[t]
	\centering
	\small
	\begin{tabular}{@{}lp{0.43\linewidth}p{0.42\linewidth}@{}}
		\toprule
		Method & Endpoint & Comment \\
		\midrule
		POST & \path|/api/rollouts| & Create a batch of rollouts. \\
		GET & \path|/api/rollouts| & List rollouts, optionally filtered by state. \\
		GET & \path|/api/rollouts/{rollout_id}| & Get one rollout. \\
		PATCH & \path|/api/rollouts/{rollout_id}| & Update rollout status. \\
		POST & \path|/api/rollouts/{rollout_id}/attempt/{attempt_id}/events| & Append an event to a rollout attempt. \\
		GET & \path|/api/rollouts/{rollout_id}/events| & Read rollout events. \\
		\midrule
		POST & \path|/api/models| & Register model endpoints. \\
		DELETE & \path|/api/models| & Remove all registered model endpoints. \\
		POST & \path|/proxy/rollout/{rollout_id}/attempt/{attempt_id}/mode/{mode}/openai/v1/chat/completions| & Forward an OpenAI-compatible model call. \\
		\bottomrule
	\end{tabular}
	\caption{API Gateway endpoints.}
	\label{tab:api-gateway-endpoints}
\end{table}

\paragraph{Rollout API.}
The trainer creates rollouts through this API. The Rollout Controller polls
queued rollouts, launches agents, and updates their status as execution
progresses. Rewards and other user-defined events are uploaded the same way.

\paragraph{Proxy API.}
This API forwards LLM calls from agent harnesses to the model endpoints the
trainer has registered. An agent harness only needs to point its
OpenAI-compatible client at the proxy. Since the proxy path embeds the
rollout ID, every call can be attributed to its rollout automatically. Each
call's prompt token IDs, response token IDs, and log probabilities are
recorded as a \texttt{model\_request} event, which the trainer later exports
for training.

This simple API fully decouples RL training from agent execution: the trainer
only creates rollouts and collects trajectories, any harness can connect by
switching its LLM endpoint to the proxy, and training and execution resources
can be provisioned independently and even run in different locations.

\subsection{Rollout Controller}

\begin{figure}[t]
	\centering
	\includegraphics[width=\linewidth]{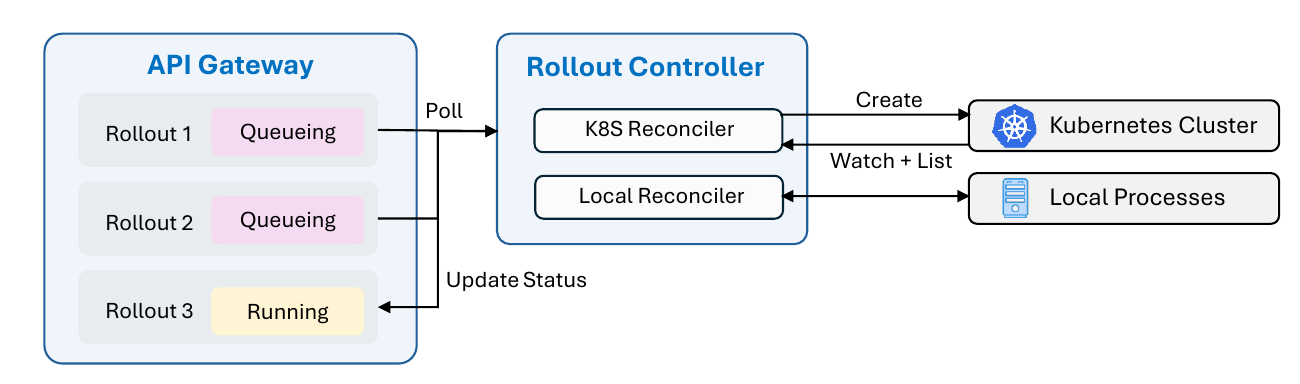}
	\caption{The Rollout Controller reconciles rollout status in the API Gateway
	with agent executions running as Kubernetes Jobs or local processes.}
	\label{fig:rollout-controller}
\end{figure}

The Rollout Controller manages agent execution on top of Kubernetes. As
shown in Figure~\ref{fig:rollout-controller}, it periodically fetches active
rollouts from the API Gateway, launches the corresponding agent tasks,
monitors them, and reports status back through the Gateway API. Its primary
backend is the K8s Reconciler, which targets an open-source Kubernetes
cluster. For debugging purposes, it also provides a Local Reconciler, which
targets a local process pool.

\paragraph{K8s Reconciler.}
For each \texttt{queuing} rollout without an existing Kubernetes Job, it
creates one from a user-provided template. It watches the Kubernetes API for
Job updates so terminal states propagate with low latency, and periodically
lists all managed Jobs to recover any watch events that were missed -- the
standard Kubernetes controller pattern.

\paragraph{Local Reconciler.}
It launches agents in a local process pool instead. Since it owns the process
handles directly, periodic polling alone is enough, and no separate watch
mechanism is needed.

\paragraph{State Consistency.}
The API Gateway's rollout status is the ground truth. The execution state
observed from Kubernetes may lag behind it because of network failures or
delayed updates. The K8s Reconciler simply retries synchronization on its
next cycle, so the two sides converge once communication resumes. This
design guarantees only best-effort eventual consistency.

\subsection{Customized Trainer}

Built on top of VERL~\cite{hybridflow}, the Customized Trainer keeps a
lightweight design while connecting the training backend to the API Gateway:
at each step it registers rollouts for the current batch, waits for them to
reach a terminal state, then retrieves their \texttt{model\_request} and
\texttt{reward} events and assembles them into training samples. It consists of the following two components.

\paragraph{Dedicated Sample Adapter.}
The adapter reflects our design choices for the challenges described in
Section~\ref{sec:challenges}.
\begin{itemize}
	\item \textbf{Sample merging.} We keep the API Gateway as simple as
	possible: it does not maintain a server-side request buffer, which keeps
	training consistent with deployment. The adapter merges two consecutive
	model requests into one training sample only when the later prompt is an
	exact token-level prefix match of the earlier request and response.
	\item \textbf{Advantage calculation.} The adapter computes baselines and
	advantages at the rollout level, which we believe is the more principled
	choice.
	\item \textbf{Loss normalization.} The adapter implements the
	rollout-level token-mean loss discussed in Section~\ref{sec:challenges},
	normalizing so every rollout carries equal weight regardless of its
	sample count.
\end{itemize}

\paragraph{Trajectory Monitoring.}
Because agentic training can produce reward hacking, runaway trajectories, or
silent failures, the trainer exposes every training and validation rollout's
input, status, model requests, rewards, token/turn statistics, and custom
events, with execution logs kept in Kubernetes, so we can inspect them
manually or with AI agents to diagnose unusual behavior.

\end{document}